\pdfoutput=1

\documentclass[11pt]{article}

\usepackage[final]{acl}

\usepackage{times}
\usepackage{latexsym}
\usepackage{booktabs} 
\usepackage{amsmath}
\usepackage{amsfonts}
\usepackage{comment}
\usepackage{mathrsfs}
\usepackage{graphicx}
\usepackage{subcaption}
\usepackage{tikz,pgfplots,pgfplotstable}
\pgfplotsset{compat=1.8}

\usepackage{diagbox}
\usepackage{multirow}
\usepackage{enumitem}
\usepackage[normalem]{ulem}

\usepackage{color, xspace}
\newcommand{\ie}{\emph{i.e.,}\xspace}
\newcommand{\eg}{\emph{e.g.,}\xspace}

\newcommand{\mname}{MC-indexing\xspace}

\usepackage[T1]{fontenc}

\usepackage[utf8]{inputenc}

\usepackage{microtype}

\usepackage{inconsolata}

\title{Multi-view Content-aware Indexing for Long Document Retrieval}

\author{Kuicai Dong$^{\dagger}$, Derrick Goh Xin Deik$^{\dagger}$, Yi Quan Lee,  Hao Zhang, \\
        {\bf Xiangyang Li, Cong Zhang, Yong Liu} \\
        Huawei Noah’s Ark Lab \\ 
        $\{$dong.kuicai; goh.xin.deik; liu.yong6$\}$@huawei.com}

\begin{document}

\maketitle
\def\thefootnote{$\dagger$}\footnotetext{These authors contributed equally}\def\thefootnote{\arabic{footnote}}

\begin{abstract} 
Long document question answering (DocQA) aims to answer questions from long documents over 10k words. They usually contain \textit{content structures} such as sections, sub-sections, and paragraph demarcations. However, the indexing methods of long documents remain under-explored, while existing systems generally employ fixed-length chunking. As they \textit{do not consider content structures}, the resultant chunks can exclude vital information or include irrelevant content. 
Motivated by this, we propose the \textbf{M}ulti-view \textbf{C}ontent-aware \textbf{indexing} (\textbf{MC-indexing}) for more effective long DocQA via (i) segment structured document into content chunks, and (ii) represent each content chunk in raw-text, keywords, and summary views. 
We highlight that MC-indexing \textit{requires neither training nor fine-tuning}. Having plug-and-play capability, it can be seamlessly integrated with any retrievers to boost their performance.
Besides, we propose a long DocQA dataset that includes not only question-answer pair, but also \textit{document structure} and \textit{answer scope}.
When compared to state-of-art chunking schemes, MC-indexing  has significantly increased the recall by \textbf{42.8\%}, \textbf{30.0\%}, \textbf{23.9\%}, and \textbf{16.3\%} via top $k=$ 1.5, 3, 5, and 10 respectively.
These improved scores are the average of 8 widely used retrievers (2 sparse and 6 dense) via extensive experiments.

\end{abstract}

\section{Introduction}
\label{sec:introduction}

\begin{figure}[ht]
\small
    \centering
    \begin{subfigure}[b]{\columnwidth}
        \rule{\linewidth}{1pt}
        \noindent \textbf{Question (a)}: \textsc{How to bake a chocolate cake?}\\[1.0mm]
        \noindent \textbf{Desired Reference Text:} You can bake a chocolate cake by following procedures: 
        1.Preparation: ... 
        2.Gather Ingredients: ...
        3.Dry Ingredients Mixture: ...
        4.Wet Ingredients Mixture: ...
        5.Combine Mixtures: ...
        6.Bake the Cake: ... (500 words) \\[1.0mm]
        \noindent \textbf{Actual Chunks Retrieved:} ... You can bake a chocolate cake by following procedures: 1.Preparation: ... (100 words) \\[-1.5mm]
        \rule{\linewidth}{0.5pt}
        \caption{The whole section (approx. 500 words) is required to answer the question. The retrieved chunk only has 100 words.}
        \label{subfig:bad_case1}
        \vspace{+0.8em}
    \end{subfigure}
    
    \begin{subfigure}[b]{\columnwidth}
        \rule{\linewidth}{0.5pt}
        \noindent \textbf{Question (b)}: \textsc{What is the hardware specifications (cpu, display, battery, etc) of Dell XPS 13?}\\[1.0mm]
        \noindent \textbf{Desired Reference Text:} ... 11th Gen Intel Core i7 processor ... a 13.4-inch FHD InfinityEdge display ... battery life ... backlit keyboard ... with Thunderbolt 4 ports ... (250 words) \\[1.0mm]
        \noindent \textbf{Actual Chunks Retrieved:} \\[0.0mm] 
        \noindent \textbf{1}. ... an 11th Gen Intel Core i7 processor ... 13.4-inch FHD InfinityEdge display ... (Content: \textbf{Dell XPS 13}, 100 words) \\[0.5mm]
        \noindent \textbf{2}. ... new M1 Pro chip ... 14-inch Liquid Retina XDR display showcases ... (Content: \textbf{MacBook Pro}, 100 words) \\[0.5mm]
        \noindent \textbf{3}. ... a powerful Intel Core M processor ... 13.3-inch 4K UHD touch display  ... (Content: \textbf{Dell XPS 12}, 100 words) \\[-1.5mm]    
        \rule{\linewidth}{1.0pt}
        \caption{The whole section (approx. 250 words) is required to answer the given question related to Dell XPS 13. Missing information (e.g, model name) leads to conflicting information.}
         \label{subfig:bad_case2}
    \end{subfigure}
    \vspace{-1.5em}
    \caption{Bad cases from fixed-length chunking due to relevant text missing and inclusion of irrelevant text.}
    \vspace{-1.5em}
    \label{fig:bad_case}
\end{figure}

Document question answering (DocQA) is a pivotal task in natural language processing (NLP) that involves responding to questions using textual documents as the reference answer scope. 
Conventional DocQA systems comprise three key components: (i) an indexer that segments the document into manageable text chunks indexed with embeddings, (ii) a retriever that identifies and fetches the most relevant chunks to the corresponding question, and (iii) a reader that digests the retrieved answer scope and generates an accurate answer.
Unlike the retriever~\cite{DBLP:journals/ftir/RobertsonZ09, karpukhin-etal-2020-dense, omar-etal-2020-colbert} and reader~\cite{nie-etal-2019-revealing, 10.5555/3495724.3496517, izacard-grave-2021-leveraging} that are vastly studied, the indexer received relatively less attention.

Existing indexing schemes \textit{overlook the importance of content structures} when dealing with long documents, as they are usually organized into chapters, sections, subsections, and paragraphs~\cite{liu-etal-2020-document, buchmann-etal-2024-document}, \ie structured.
The widely used fixed-length chunking strategy can easily break the contextual relevance between text chunks for long documents. Such chunking errors can be further aggravated by the retriever and the reader.
Moreover, determining the boundary between chunks can be tricky, requiring delicate design to prevent contextual coherence disruption.
Ideally, each chunk should represent a coherent and content-relevant textual span. 
Otherwise, it can lead to the exclusion of relevant information or the inclusion of irrelevant text, as exemplified in Figure~\ref{fig:bad_case}.
Our empirical study on fixed-length chunking reveals that setting the chunk length to $100$ results in over $70\%$ of long answers/supporting evidence being truncated, \ie incomplete. Such incompleteness still exists at $45\%$, despite an increase of chunk length to $200$.\footnote{More statistics of chunking errors are in Section~\ref{ssec:chunking}.}

\begin{figure} [t]
    \centering
    \begin{subfigure}[b]{0.45\columnwidth}
         \includegraphics[trim={0.0cm, 0.0cm, 0, 0.0cm},clip, width=\linewidth]{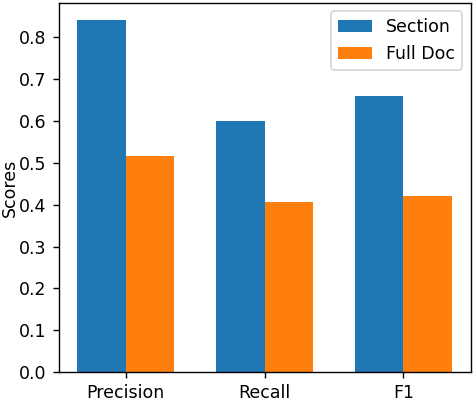}
         \caption{Medium-size LLM on document (12-15k tokens)}
         \label{subfig:med_llm}
    \end{subfigure}
    ~
    \begin{subfigure}[b]{0.46\columnwidth}
         \includegraphics[trim={0.0cm, 0, 0, 0.0cm},clip, width=\linewidth]{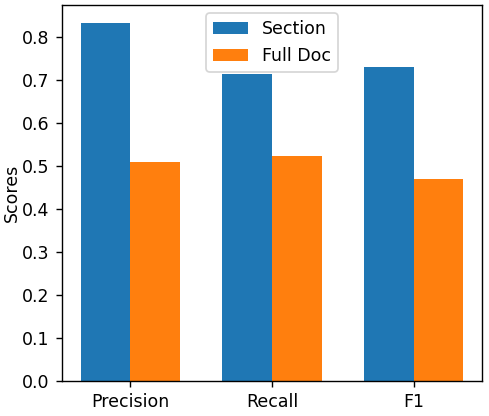}
         \caption{Large-size LLM on document (28-30k tokens)}
         \label{subfig:large_llm}
    \end{subfigure}
    \vspace{-0.5em}
    \caption{Commercial LLM on Span-QA retrieval using full document vs dedicated section.}
    \vspace{-1em}
    \label{fig:llm_long_context}
\end{figure}

Recently, there is a growing interest in utilizing Large Language Models (LLMs) for QA tasks~\cite{chen2023walking, sarthi2024raptor}.
However, the notorious token limit constraint puts a barricade for applying LLM to long documents.
For instance, LLaMA~\cite{touvron2023llama}, LLaMA 2~\cite{touvron2023llama2}, and Mistral~\cite{jiang2023mistral} are limited to 2k, 4k, and 8k tokens, respectively, which is insufficient for long documents with more than 10k words.
Even advanced commercial LLMs struggle to effectively process and understand such long documents. 
As depicted in Figure~\ref{fig:llm_long_context}, our research indicates that the performance of advanced commercial LLM deteriorates substantially when applied to span-based QA on long documents (12k-30k tokens) as compared to a dedicated section (370 tokens in average).\footnote{More results for long DocQA are in Appendix~\ref{ssec:gpt_long_context}.} 
Furthermore, \citet{liu2023lost} indicate that LLMs face difficulties in retaining and referencing information from earlier portions of long documents.

To mitigate the aforementioned challenges, we present a new approach \textbf{M}ulti-view \textbf{C}ontent-aware \textbf{Indexing}, termed \textbf{\mname}, which allows for a more effective retrieval of long documents.
Our method involves content-aware chunking of structured long documents, whereby, instead of employing na\"ive fixed-length chunking, the document is segmented into section chunks. Each of these section chunks is then indexed in three different views, representing each chunk with raw-text, a list of keywords, and a summary.
This multi-view representation significantly enhances the semantic richness of each chunk. For retrieval, we aggregate the top relevant chunks based on multi-views.
Note that the entire process of \mname is unsupervised. 
We leverage on the strength of existing retrievers for the embedding generation of raw-text, keyword, and summary views.
We leverage on the answer composition capability of LLM for answer generation, based on retrieved chunks.
To our best knowledge, existing DocQA datasets do not provide content structure. Hence, we transform an existing long documents dataset, namely WikiWeb2M~\cite{burns2023wikiweb2m}, into a QA dataset, by adding annotations to the documents.
In addition, we complement Natural Questions dataset~\cite{kwiatkowski-etal-2019-natural} with content structure, and filter only long documents for our experiment.
Distinct from other QA datasets, our documents are longer (averaging at 15k tokens) and contain detailed content structure. 
Our contributions are in fourfold:
\begin{itemize}[leftmargin=*, itemsep=-0.4em, topsep=-0.0em]
    \item We propose a long document QA dataset annotated with question-answer pair, document content structure, and scope of answer.
    \item We propose \textbf{M}ulti-view \textbf{C}ontent-aware \textbf{indexing} (\textbf{\mname}), that can (i) segment the long documents according to their content structures, and (ii) represent each chunk in three views, \ie raw-text, keywords, and summary.
    \item \mname requires neither training nor fine-tuning, and can seamlessly act as a plug-and-play indexer to enhance any existing retrievers. 
    \item Through extensive experiments and analysis, we demonstrate that \mname can significantly improve retrieval performance of \textbf{eight} commonly-used retrievers (2 sparse and 6 dense) on two long DocQA datasets.
\end{itemize}
\section{Related Work}
\label{sec:related}

\paragraph{Retrieval Methods.}

Current approaches to content retrieval are primarily classified into sparse and dense retrieval. There are two widely-used sparse retrieval methods, namely TF-IDF~\cite{TF-IDF} and BM25~\cite{bm25}. TF-IDF calculates the relevance of a word to a document in the corpus by multiplying the word frequency with the inverse document frequency. BM25 is an advancement of TF-IDF that introduces nonlinear word frequency saturation and length normalization to improve retrieval accuracy. 

Recently, dense retrieval methods have shown promising results, by encoding content into high-dimensional representations. DPR~\cite{karpukhin-etal-2020-dense} is the pioneering work of dense vector representations for QA tasks. Similarly, ColBERT~\cite{khattab-etal-2020-colbert} introduces an efficient question-document interaction model, enhancing retrieval accuracy by allowing fine-grained term matching. 
Contriever~\cite{izacard-etal-2022-contriever} further leverages contrastive learning to improve content dense encoding.
E5~\cite{wang2022-e5} and BGE~\cite{xiao2023-bge} propose novel training and data preparation techniques to enhance retrieval performance, \eg consistency-filtering of noisy web data in E5 and the usage of RetroMAE~\cite{xiao2023-retromae} pre-training paradigm in BGE.
Moreover, GTE~\cite{li2023-gte} integrates graph-based techniques to enhance dense embedding. 

In summary, these systems focus on how to retrieve relevant chunks, but neglecting how text content is chunked.
In contrast, \mname  can utilize the strengths of existing retrievers, and further improve their retrieval performance.

\paragraph{Chunking Methods.}
Chunking is a crucial step in either QA or Retrieval-Augmented Generation (RAG). When dealing with ultra-long text documents, chunk optimization involves breaking the document into smaller chunks.
In practice, fixed-length chunking is a commonly used method that is easy to be implemented. It chunks text at a fixed length, \eg 200 words. 
Sentence chunking involves dividing textual content based on sentences.
Recursive chunking employs various delimiters, such as paragraph separators, newline characters, or spaces, to recursively segment the text.
However, these methods often fail to preserve semantic integrity of critical content.
In contrast, content-aware chunking (Section~\ref{ssec:content-chunk}) chunk the text by the smallest subdivision according to the document's content structure.
This ensures each chunk to be semantically coherent, thus reducing chunking error.

\paragraph{Long Document Retrieval.}
Traditional retrieval methods such as BM25 and DPR only retrieve short consecutive chunks from the retrieval corpus, limiting the overall understanding of the context of long documents. To overcome this drawback, several methods focusing on long document retrieval have been proposed. \citet{nie-etal-2022-capturing} propose a compressive graph selector network to select question-related chunks from the long document and then use the selected short chunks for answer generation.
AttenWalker~\cite{nie-etal-2023-attenwalker} addresses the task of incorporating long-range information by employing a meticulously crafted answer generator.
\citet{chen2023walking} convert the long document into a tree of summary nodes. Upon receiving a question, LLM navigates this tree to find relevant summaries until sufficient information is gathered.
\citet{sarthi2024raptor} utilize recursive embedding, clustering, and summarizing chunks of text to build a tree with different levels of summarization. 
However, existing methods only consider the retrieval of long documents from one view, limiting the semantic completeness and coherence.
\section{Methodology}
\label{sec:method}

\begin{figure*}
    \centering
    \begin{subfigure}[b]{1.65\columnwidth}
         \includegraphics[width=1.0\linewidth]{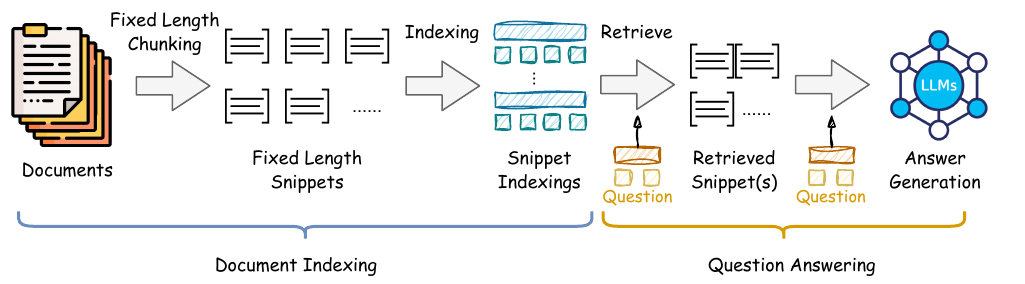}
         \vspace{-2em}
         \caption{Conventional DocQA system: document -> fixed length snippets -> retrieved snippets -> answer}
         \vspace{+0.5em}
         \label{subfig:conventiaon_qa}
    \end{subfigure}
    \begin{subfigure}[b]{2.0\columnwidth}
         \includegraphics[width=1.0\linewidth]{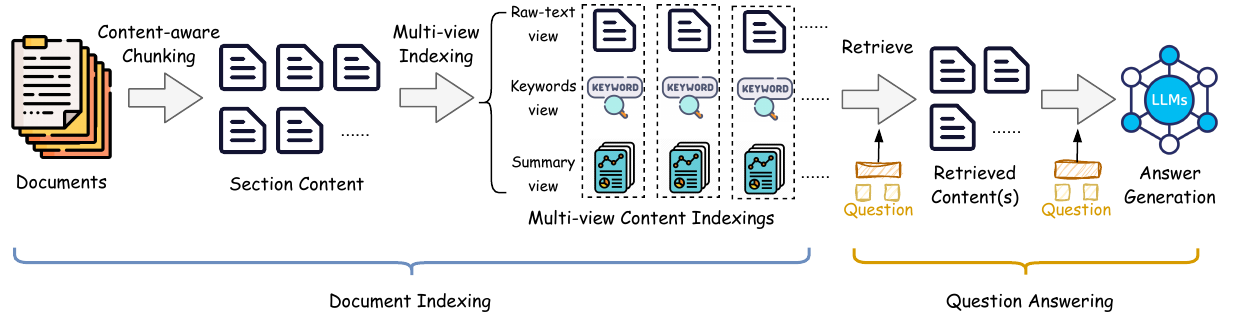}
         \vspace{-2em}
         \caption{Our method: document ->  section content -> multi-view content indexing -> retrieved sections -> answer}
         \label{subfig:llm_qa}
    \end{subfigure}
    \vspace{-0.5em}
    \caption{Comparison between conventional fixed length chunking and content-aware LLM QA systems. }
    \vspace{-1em}
    
    \label{fig:model_comparison}
\end{figure*}

\subsection{Content-aware Chunking}
\label{ssec:content-chunk}

We elaborate how Content-Aware chunking is performed in order to obtain \emph{section chunks}. 
Given a piece of structured document (\eg Markdown, Latex, Wikipedia Articles), we first extract the table of contents of the document (or header information, in the event where the table of content is not readily available).
Upon acquiring this information, we identify the smallest division in the document, such as a section, subsection, or sub-subsection, depending on the structure of the content. 
It is reasonable to assume that these smallest divisions function as atomic, coherent semantic units within the document.
The text present in each smallest division is the desired \emph{section chunk}.

Chunking text based on the smallest division, as opposed to fixed length chunking, ensures that information in each chunk cannot contain information across two different sections. 
Most importantly, we preserve the semantic integrity during the chunking process, leading to each section chunk to be an atomic and coherent semantic unit.
Note that different sections may have a hierarchical relationship between them. We ignore them for now and assume a flat structure between different chunks.

\subsection{Multi-View Indexing and Retrieval}
\label{ssec:multi-view}

Most dense retrieval methods primarily uses raw text from each chunk to determine the relevancy of each chunk with respect to a given question. 
It works well when using fixed-length chunk to ensure segments do not exceed the dense retrieval models' token limits. 
However, our content-aware chunking approach, which divides documents into their smallest meaningful divisions, often results in chunks larger than the models' maximum token capacity. As indicated in Table~\ref{tab:doc_stats}, Wiki-NQ dataset has averaging 500 tokens per section.
This can lead to truncation and information loss during retrieval, undermining the effectiveness of the method.

To mitigate the aforementioned issue, it is necessary to introduce more concise representations or \emph{views} of section chunks to improve retrieval efficiency. 
Beyond the \emph{\textbf{raw text}} view, which consists solely of the section's text, we also introduce the \emph{\textbf{summary}} view and the \emph{\textbf{keyword}} view.  

The \emph{\textbf{summary}} view, as the name suggests, represents each section chunk with a succinct summary.
It captures the key information of each section. 
The summary is generated by generative LLM using the prompt given in Figure~\ref{fig:promp_summary}. 
This ensures that the summary fits within the dense retrieval model's maximum input limit during retrieval.

To compensate for the potential omission of critical details in the generated summaries, we introduce a \emph{\textbf{keyword}} view. 
This view characterizes each section chunk by a list of essential keywords, including significant concepts, entities, and terms from the section. We employ a prompt outlined in Figure~\ref{fig:promp_keywords}, enabling generative LLM to identify and list the keywords associated with each section.

Finally, we describe the procedure for utilizing multi-view indexing to retrieve top-$k$ relevant sections with respect to a given question.
For each of the views, \eg raw-text, summary, keywords, we simply rank the sections using each view to first retrieve the top-$k^{'}$ results. 
Setting $k^{'} \approx 2k/3$ works since empirically we expect on average a total of $3k^{'}/2$ unique results after deduplication. 
Thereafter we feed the set of raw text to LLM for answer generation using prompt in Figure~\ref{fig:promp_ac}.  
Note that \mname is independent of retriever selection.
It can be used to augment any retriever by supplying top-ranked chunks across different views.
As a plug-and-play boost for all existing retrievers, \mname requires no additional training or fine-tuning to integrate effectively.

\section{Dataset Construction}
\label{sec:dataset}
In our work, we focus on long and structured document, thus we collect dataset corpus based on the following two factors.
\textbf{(1) Presence of structured information:} The content of long documents is usually divided into multiple sections. For example, a research paper is organized into various sections such as Abstract, Introduction, Methodology and Conclusion. Structured documents have explicitly labelled sections along their corresponding text. 
Most of the existing QA datasets (\eg SQuAD~\cite{rajpurkar-etal-2016-squad}, TriviaQA~\cite{joshi-etal-2017-triviaqa}, MS MARCO \cite{bajaj2018ms}) focus on retrieval of relevant passages.
Thus the original organizational structure of the source documents are usually omitted. 
However, for the purpose of \mname, it is necessary to preserve structure information.
\textbf{(2) Sufficiently Long Document:} Our method leverages using the smallest division as a section chunk. In particular, if a document is too short, a document would be divided into only a few section chunks under Content-aware Chunking. To avoid these trivial cases, we only select documents with at least 10k words, as the number of sections correlates positively with document length. 

According to these criteria, we select Wikipedia Webpage 2M (WikiWeb2M) \cite{burns2023wikiweb2m} and Natural Questions (NQ) \cite{kwiatkowski-etal-2019-natural} datasets. We discuss dataset processing and annotations on these datasets in finer detail.

\subsection{Wikipedia Webpage 2M (WikiWeb2M)} \label{ssec:data_wiki2m}
WikiWeb2M is designed for multimodal webpage understanding rather than QA.
The dataset stores individual sections within each Wikipedia article.
Thus, on top of the structured information, we annotate additional question-answer pairs and their answer scope.
We utilize advanced commercial LLM to construct questions for selected articles (over 10k tokens) in WikiWeb2M. 
To ensure that the questions rely on long answer scope span, we define the 8 types of questions.\footnote{See Appendix ~\ref{ssec:q_type} for more details of question type definition and statistics.}
For each section given, we request LLM (using prompt shown in Figure \ref{fig:prompt_query}) to generate (i) three questions, (ii) the corresponding answers to the each question, and (iii) the answer scope for each answer.
We then evaluate the retrieval efficiency and answer quality of \mname by utilizing the constructed data. 

Using this approach we have generated questions for 83,625 sections from 3,365 documents. For evaluation, in order to demonstrate the effectiveness of our method in long DocQA, we only use questions generated from documents with 28k to 30k tokens, resulting in 30 documents for evaluation. 
The remaining questions not used in evaluation are intended for training / fine-tuning, if needed. Note that our proposed method (in Section \ref{sec:method}) does not require any form of fine-tuning.

\subsection{Natural Questions (NQ)} \label{ssec:data_wikinq}
The NQ dataset provides rendered HTML of Wikipedia articles alongside the questions and answer scope.
By parsing the rendered HTML, we are able to extract the section name and the corresponding texts in each section of the document. 
We augment the NQ dataset with our extracted structured information.
We omit sections such as `See Also', `Notes', and `References', which refer as references for the main content, to reduce noise during retrieval.
We follow NQ's train/dev split setting in our work.
However, we only retain the question where its corresponding document has more than 10k tokens.  
In addition, we filter out queries where the annotated answer scope spans across multiple sections. For dev set, there exists multiple annotations. We only retain questions where all annotations reside within the same section.
After filtering, we obtain 36,829 and 586 question-article pairs for train/dev respectively. 
Again, we emphasise that our approach does not require fine-tuning and solely utilises the dev-set.

\begin{table}[t] 
\small
    \centering
  \begin{tabular}{c|cccc}
    \toprule
    \multirow{2}{*}{Statistics} &  \multicolumn{2}{c}{NQ} & \multicolumn{2}{c}{WikiWeb2M} \\
    & Dev & Train & Dev & Train \\
    \midrule
    questions & 586 & 36.8k & 3027 & 82.6k\\
    sections & 34.1 & 33.2 & 75.0 & 42.7\\
    tokens/doc & 17.4k & 17.4k & 28.1k & 15.2k \\
    tokens/sec & 510 & 525 & 375 & 356 \\
    tokens/ans & 827 & 581 & 109 & 104\\
  \bottomrule
  \end{tabular}
\vspace{-0.6em}
\caption{Document statistics for NQ and WikiWeb2M.}
\vspace{-2em}
\label{tab:doc_stats}
\end{table}
\section{Experiment}
\label{sec:experiment}

\begin{table*}[t] 
\small
    \centering
    \resizebox{\linewidth}{!}{%
  \begin{tabular}{ll|p{0.36cm}p{0.36cm}p{0.36cm}p{0.45cm}|p{0.36cm}p{0.36cm}p{0.36cm}p{0.36cm}p{0.36cm}p{0.36cm}p{0.36cm}p{0.36cm}p{0.3cm}p{0.3cm}p{0.3cm}p{0.4cm}|p{0.36cm}}
    \toprule
    \multicolumn{2}{c}{\multirow{3}{*}{Chunking Scheme}}
    &  \multicolumn{4}{|c|}{Sparse Retriever} & \multicolumn{12}{c|}{Dense Retriever} & \multirow{3}{*}{Ave} \\ 
    \cmidrule{3-18}
     & & \multicolumn{2}{c}{TF-IDF} & \multicolumn{2}{c|}{BM25} &  \multicolumn{2}{c}{DPR} & \multicolumn{2}{p{0.7cm}}{ColBERT} & \multicolumn{2}{p{0.7cm}}{Contirever} & \multicolumn{2}{c}{E5} & \multicolumn{2}{c}{BGE} & \multicolumn{2}{c|}{GTE} \\
     & & 2M & NQ & 2M & NQ & 2M & NQ & 2M & NQ & 2M & NQ & 2M & NQ & 2M & NQ & 2M & NQ \\
    \midrule
    \parbox[t]{1mm}{\multirow{10}{*}{\rotatebox[origin=c]{90}{Top $k=1.5$}}} &
    FLC: 100 tokens  & 47.8 & 14.6 & 45.8 & 7.8 & 35.3 & 25.1 & 54.2 & 27.4 & 54.2 & 22.9 & 57.7 & 33.0 & 55.8 & 27.9 & 56.3 & 29.8 & 37.2 \\
    &\quad\quad-content  & 48.0 & 14.9 & 48.6 & 8.2 & 38.7 & 29.6 & 61.0 & 32.4 & 60.6 & 28.1 & 63.4 & 37.6 & 62.7 & 34.7 & 62.7 & 34.9 & 41.6 \\
    & FLC: 200 tokens  & 51.1 & 19.4 & 56.1 & 11.7 & 40.6 & 35.7 & 62.0 & 37.1 & 61.9 & 29.8 & 67.0 & 41.9 & 63.2 & 37.3 & 63.7 & 38.1 & 44.8 \\
    & \quad\quad-content  & 58.9 & 22.0 & 60.6 & 14.2 & 45.5 & 40.8 & 71.0 & 42.7 & 70.1 & 36.3 & 73.6 & 46.4 & 71.9 & 42.1 & 72.4 & 44.7 & 50.8 \\
    & FLC: 300 tokens  & 60.9 & 20.8 & 61.6 & 13.9 & 41.5 & 41.3 & 64.0 & 37.5 & 64.4 & 35.0 & 68.1 & 47.9 & 64.6 & 41.1 & 65.1 & 41.8 & 48.1 \\
    & \quad\quad-content  & 64.5 & 24.6 & 66.3 & 15.0 & 48.8 & 43.3 & 73.0 & 45.5 & 73.5 & 38.1 & 75.9 & 48.8 & 75.1 & 45.5 & 75.7 & 49.5 & 53.9\\
    
    & Content: raw-text  & 59.0 & 22.5 & 66.7 & 19.6 & 49.0 & 39.6 & 67.1 & 43.2 & 72.1 & 34.5 & 76.3 & 43.5 & 72.7 & 45.9 & 74.0 & 47.8 & 52.1 \\
    & Content: keyword  & 47.4 & 16.7 & 57.8 & 12.8 & 46.5 & 31.3 & 69.2 & 38.9 & 67.0 & 30.4 & 70.0 & 44.2 & 65.8 & 39.8 & 68.3 & 41.0 & 46.7 \\
    & Content: summary  & 66.2 & 24.4 & 72.2 & 17.6 & 54.3 & 43.3 & 74.0 & 42.7 & 72.8 & 37.0 & 73.3 & 53.2 & 71.8 & 47.4 & 73.3 & 45.6 & 54.3 \\
    & \mname & \textbf{79.2} & \textbf{40.9} & \textbf{83.7} & \textbf{36.9} & \textbf{67.7} & \textbf{58.4} & \textbf{85.1} & \textbf{62.3} & \textbf{83.8} & \textbf{52.2} & \textbf{87.0} & \textbf{69.6} & \textbf{83.7} & \textbf{63.1} & \textbf{84.0} & \textbf{62.3} & \textbf{68.7}\\
    \midrule
    \midrule
    
    \parbox[t]{1mm}{\multirow{10}{*}{\rotatebox[origin=c]{90}{Top $k=3$}}} &
    FLC: 100 tokens  & 58.3 & 21.2 & 58.7 & 12.9 & 46.9 & 35.4 & 64.4 & 39.2 & 65.0 & 35.2 & 69.5 & 46.3 & 69.4 & 41.1 & 69.5 & 43.0 & 48.5 \\
    & \quad\quad-content  & 62.3 & 22.9 & 61.0 & 14.0 & 53.4 & 42.4 & 71.6 & 44.2 & 73.3 & 37.4 & 77.0 & 49.0 & 74.9 & 46.4 & 76.8 & 47.7 & 53.4 \\
    & FLC: 200 tokens  & 67.7 & 30.2 & 70.2 & 21.9 & 55.0 & 48.7 & 70.9 & 50.8 & 73.5 & 43.6 & 77.8 & 56.7 &75.7 &  52.9 & 77.5 & 54.2 & 58.0 \\
    & \quad\quad-content  & 72.9 & 33.3 & 71.7 & 23.0 & 51.6 & 55.3 & 81.8 & 55.3 & 83.4 & 47.5 & 84.3 & 62.3 &82.7 &  56.4 & 85.2 & 58.7 & 62.8 \\
    & FLC: 300 tokens  & 70.7 & 32.3 & 74.9 & 23.7 & 58.4 & 54.4 & 73.8 & 50.0 & 75.6 & 51.7 & 81.2 & 62.1 & 77.7 &  57.6 & 78.2 & 59.2 & 61.3 \\
    & \quad\quad-content  & 76.2 & 36.9 & 76.4 & 24.8 & 61.8 & 58.4 & 82.5 & 59.6 & 85.0 & 54.4 & 86.9 & 65.0 &85.5 &  61.0 & 87.1 & 64.8 & 66.6\\
    & Content: raw-text  & 75.2 & 46.8 & 81.4 & 41.6 & 66.5 & 69.5 & 80.0 & 68.9 & 86.1 & 62.6 & 88.1 & 77.3 & 85.6 & 73.9 & 86.4 & 74.4 & 72.8 \\
    & Content: keyword  & 69.5 & 39.9 & 73.8 & 30.7 & 64.9 & 59.7 & 84.2 & 65.5 & 82.5 & 63.3 & 83.6 & 75.6 & 83.3 & 70.1 & 84.5 & 70.3 & 68.8 \\
    & Content: summary  & 83.1 & 51.9 & 86.1 & 39.1 & 71.1 & 72.4 & 86.8 & 71.1 & 86.6 & 64.5 & 88.1 & 81.6 & 86.9 & 76.9 & 87.3 & 76.3 & 75.6 \\
    & \mname & \textbf{86.6} & \textbf{54.1} & \textbf{89.3} & \textbf{47.6} & \textbf{77.2} & \textbf{75.1} & \textbf{91.0} & \textbf{77.1} & \textbf{90.5} & \textbf{70.8} & \textbf{92.8} & \textbf{85.3} & \textbf{90.6} & \textbf{78.8} & \textbf{90.8} & \textbf{77.8} & \textbf{79.7}\\
    \midrule
    \midrule

    \parbox[t]{1mm}{\multirow{10}{*}{\rotatebox[origin=c]{90}{Top $k=5$}}} & 
    FLC: 100 tokens  & 65.5 & 28.4 & 65.2 & 19.2 & 54.8 & 45.4 & 70.6 & 46.7 & 70.9 & 43.3 & 77.7 & 55.2 & 75.8 & 50.8 & 76.8 & 52.0 & 56.1\\
    & \quad\quad-content  & 67.9 & 30.3 & 67.0 & 20.0 & 61.3 & 51.9 & 77.5 & 52.8 & 78.7 & 46.4 & 82.3 & 58.2 & 80.5 & 54.3 & 82.6 & 56.5 & 60.5 \\
    & FLC: 200 tokens  & 74.1 & 39.2 & 77.2 & 30.1 & 64.9 & 60.2 & 76.1 & 59.5 & 78.9 & 54.0 & 83.6 & 66.3 & 81.6 & 61.6 & 82.4 & 63.9 & 65.9 \\
    & \quad\quad-content  & 77.8 & 41.2 & 77.2 & 31.0 & 59.5 & 62.0 & 85.9 & 63.9 & 87.6 & 59.8 & 89.1 & 71.9 & 87.1 & 65.6 & 89.5 & 67.4 & 69.8 \\
    & FLC: 300 tokens  & 76.7 & 42.5 & 80.8 & 34.9 & 65.7 & 66.8 & 78.8 & 60.3 & 81.9 & 62.8 & 85.9 & 73.1 & 83.1 & 68.6 & 84.1 & 70.0 & 69.8 \\
    & \quad\quad-content  & 80.3 & 47.2 & 81.1 & 35.6 & 69.4 & 67.1 & 87.1 & 68.4 & 89.0 & 64.2 & 90.4 & 75.9 & 89.5 &  69.1 & 91.2 & 71.9 & 73.6 \\
    & Content: raw-text & 80.0 & 63.5 & 85.3 & 53.8 & 74.2 & 80.7 & 84.5 & 78.2 & 90.2 & 74.2 & 91.3 & 87.9 & 89.2 & 82.6 & 89.7 &  84.1 & 80.6 \\
    & Content: keyword  & 76.5 & 53.8 & 80.2 & 43.3 & 73.0 & 75.1 & 89.0 & 76.6 & 87.5 & 75.8 & 87.8 & 85.8 & 87.8 & 82.8 & 88.9 & 82.0 & 77.9 \\ 
    & Content: summary  & 88.1 & 66.5 & 89.5 & 51.9 & 78.2 & 84.8 & 90.7 & 81.9 & 90.8 & 78.1 & 91.7 & 90.9 & 90.7 & 86.4 & 91.2 & 86.5 & 83.6 \\
    & \mname & \textbf{90.5} & \textbf{67.6} & \textbf{93.6} & \textbf{60.1} & \textbf{81.9} & \textbf{87.5} & \textbf{93.4} & \textbf{85.2} & \textbf{92.8} &  \textbf{82.1} & \textbf{94.5} & \textbf{91.8} & \textbf{93.0} & \textbf{89.2} & \textbf{93.1} & \textbf{88.0} & \textbf{86.5}\\
    
    \midrule
    \midrule

    \parbox[t]{1mm}{\multirow{10}{*}{\rotatebox[origin=c]{90}{Top $k=10$}}} &
    FLC: 100 tokens & 73.3 & 38.8 & 73.0 & 29.2 & 65.7 & 60.9 & 77.8 & 60.3 & 80.0 & 55.9 & 83.8 & 68.6 & 83.2 & 63.6 & 83.9 & 64.8 & 66.4 \\
    & \quad\quad-content & 74.4 & 41.3 & 73.5 & 30.9 & 72.6 & 62.8 & 83.2 & 62.6 & 84.6 & 59.2 & 87.1 & 70.5 & 86.1 & 65.3 & 88.0 & 65.6 & 69.2 \\
    & FLC: 200 tokens & 81.1 & 52.4 & 83.5 & 44.2 & 74.9 & 73.8 & 82.5 & 70.8 & 85.5 & 69.8 & 88.4 & 78.7 & 88.2 & 75.2 & 88.5 & 75.8 & 75.8 \\
    & \quad\quad-content & 82.7 & 58.3 & 82.4 & 42.4 & 68.6 & 74.4 & 90.7 & 75.2 & 90.6 & 72.4 & 92.9 & 80.1 & 91.3 & 77.0 & 93.4 & 77.4 & 78.1 \\
    & FLC: 300 tokens & 82.7 & 60.8 & 86.9 & 52.1 & 75.6 & 79.7 & 85.7 & 75.8 & 87.9 & 77.6 & 89.9 & 85.1 & 89.0 & 83.3 & 89.9 & 81.1 & 80.2 \\
    & \quad\quad-content & 85.2 & 64.9 & 85.4 & 51.8 & 78.5 & 78.0 & 91.8 & 79.2 & 93.0 & 76.6 & 93.7 & 84.4 & 92.8 & 82.0 & 95.1 & 82.0 & 82.2 \\
    & Content: raw-text & 85.3 & 82.4 & 89.3 & 74.2 & 83.5 & 89.9 & 90.2 & 90.6 & 93.6 & 88.7 & 94.3 & 96.2 & 92.6 & 93.7 & 93.7 & 93.0 & 89.5 \\
    & Content: keyword & 84.5 & 76.6 & 86.8 & 67.2 & 82.3 & 89.8 & 92.9 & 90.8 & 91.9 & 89.2 & 93.0 & 94.4 & 93.0 & 92.2 & 93.7 & 92.5 & 88.2 \\
    & Content: summary & 92.9 & 84.5 & 93.3 & 76.8 & 86.9 & 94.2 & 94.3 & 92.2 & 94.4 & 90.9 & 95.2 & 96.4 & 94.1 & 94.5 & 94.6 & 94.5 & 91.8 \\
    & \mname & \textbf{94.5} & \textbf{85.7} & \textbf{95.3} & \textbf{78.2} & \textbf{88.8} & \textbf{95.0} & \textbf{96.0} & \textbf{94.8} & \textbf{95.8} & \textbf{92.7} & \textbf{96.5} & \textbf{97.2} & \textbf{95.3} & \textbf{95.4} & \textbf{96.0} & \textbf{95.4} & \textbf{93.3}\\
    
  \bottomrule
  \end{tabular}
  }
\vspace{-1em}
\caption{Main results: recall of ground truth span. Last column is the average scores of all columns.}
\vspace{-1em}
\label{tab:main}
\end{table*}

\subsection{Baseline Systems}
When conducting our experiments, we primarily employ the Fixed-Length Chunking (FLC) scheme as our baseline.
We firstly segment the document into individual sentences to avoid the first and last sentence in each chunk being truncated. We merge the sentences accordingly into chunks, with approximately 100, 200 or 300 tokens.

We also evaluate on the capability of content awareness in boosting FLC. We first segment the document into section chunks, and further apply FLC on each section. Hence, a section may have multiple chunks but each chunk can only be associated with a section. This is denoted as FLC-content. We also evaluate the variations of our \mname on multi-view as mentioned in Section ~\ref{ssec:multi-view}. In comparison to raw text, these views offer a different perspective on how to approach and analyze data.

We apply \mname and all baselines on 2 sparse (TF-IDF and BM25) and 6 dense (DPR, ColBERT, Contriever, E5, BGE, and GTE) retrievers in Section~\ref{sec:related}.
Implementation details of these retrievers can be found in Appendix~\ref{appendix:implementation_retriever}.

\subsection{Evaluation Metrics} \label{ssec:evaluation_metric}
We evaluate the performance of \mname and other baselines based on (i) recall of retrieval and (ii) quality of answer generation.

\noindent
\textbf{Recall of Retrieval.} 
The retriever scores each chunk in the document based on its relevance to the question, and returns the top $k$ chunks with the highest scores.
We define recall as the proportion of the ground truth answer scope that is successfully retrieved by retriever. 
For instance, if each of three retrieved chunks overlaps with 10\%, 50\% and 0\% of the ground truth answer scope, the recall is the sum of all individual scores to be 0.6.
The recall gives us a clear indication of how effective our chunking strategy has boosted the retriever.

Due to the fact \mname combines the results from three views, we reduce the number of chunks retrieved from each view to have a fair comparison with single-view baselines.
Specifically, when comparing with top $k=3$ single-view baselines, \mname will only retrieve top $k=1$ or $2$ from each view. By combining the chunks from each view and remove overlapping ones, \mname manages to retrieve an approximate of 3 chunks in total. 
Similarly for top $k=5$, our method retrieves only 3 chunks form each view. For top $k=10$, our method retrieves 6 or 7 chunks from each view. 
To evaluate the performance of our method in greedy ranking, our method retrieves exactly 1 chunk from each view, while other baselines retrieves 1.5 chunks in average. This is achieved by retrieving 1 chunk for half of the questions and 2 chunks for the other half.

\noindent
\textbf{Answer Generation.}
As the final goal of DocQA is to generate accurate answer, it is essential for us to evaluate the quality of final answer based on retrieved chunks.
We evaluate the answers via pairwise evaluation using advanced commercial LLM as evaluator. Specifically, we provide prompt for LLM (see Figure~\ref{fig:promp_score}) to score each answer. To avoid any positional bias, which may cause the LLM model to favor the initial displayed answer,
we switch answer positions in two evaluation rounds. The winning answer is determined based on scores in two rounds.

For \textit{Score-based evaluation}, each answer's scores from the two rounds are combined. The answer with higher overall score is the winner. The result is a tie if both answers have same score.
For \textit{Round-based evaluation}, the scores from each round are compared, and the winner of each round is determined by the higher score. The overall winner is the one that wins both rounds. In cases where each answer wins a round, or answers tie in both rounds, the result is marked as a tie.

\begin{table}[t] 
\small
    \centering
    \resizebox{\linewidth}{!}{%
  \begin{tabular}{ll|cccc|c}
    \toprule
    \multicolumn{2}{c|}{\multirow{1}{*}{Chunk Scheme}}
    &  Top1.5 & Top3 & Top5 & Top10 & $\Delta$ \\
    \midrule
    \parbox[t]{0.0mm}{\multirow{4}{*}{\rotatebox[origin=c]{90}{TF-IDF}}} 
    & \mname         & \textbf{79.2} & \textbf{86.6} & \textbf{90.5} & \textbf{94.5} & - \\
    & - w/o raw text & 71.2 & 82.6 & 87.4 & 93.3 & -4.1\\
    & - w/o keyword  & 76.8 & 85.6 & 89.1 & 93.8 & -1.4\\
    & - w/o summary  & 68.2 & 77.8 & 82.1 & 87.9 & -8.7\\
    \midrule

    \parbox[t]{0.0mm}{\multirow{4}{*}{\rotatebox[origin=c]{90}{BM25}}} 
    & \mname          & \textbf{83.7} & \textbf{89.3} & \textbf{93.6} & \textbf{95.3} & - \\
    & - w/o raw text  & 78.2 & 85.9 & 91.0 & 93.8 & -3.2\\
    & - w/o keyword   & 81.6 & 87.8 & 92.1 & 94.0 & -1.6\\
    & - w/o summary   & 74.9 & 83.8 & 88.4 & 91.5 & -5.8\\
    \midrule

    \parbox[t]{0.0mm}{\multirow{4}{*}{\rotatebox[origin=c]{90}{DPR}}} 
    & \mname          & \textbf{67.7} & \textbf{77.2} & \textbf{81.9} & \textbf{88.8} & - \\
    & - w/o raw text  & 61.3 & 72.0 & 77.6 & 86.1 & -4.7\\
    & - w/o keyword   & 63.6 & 73.9 & 79.2 & 86.7 & -3.0\\
    & - w/o summary   & 59.3 & 69.9 & 75.6 & 84.2 & -6.7\\
    \midrule

    \parbox[t]{0.0mm}{\multirow{4}{*}{\rotatebox[origin=c]{90}{ColBERT}}} 
    & \mname          & \textbf{85.1} & \textbf{91.0} & \textbf{93.4} & \textbf{96.0} & - \\
    & - w/o raw text  & 82.3 & 89.5 & 91.8 & 95.3 & -1.7 \\
    & - w/o keyword   & 82.0 & 88.6 & 91.3 & 94.4 & -2.3 \\
    & - w/o summary   & 78.4 & 86.3 & 90.1 & 94.1 & -4.2 \\
    \midrule

    \parbox[t]{0.0mm}{\multirow{4}{*}{\rotatebox[origin=c]{90}{Contriever}}} 
    & \mname          & \textbf{83.8} & \textbf{90.5} & \textbf{92.8} & \textbf{95.8} & - \\
    & - w/o raw text  & 79.1 & 87.4 & 90.4 & 94.7 & -2.8 \\
    & - w/o keyword   & 81.5 & 89.0 & 91.5 & 95.0 & -1.5 \\
    & - w/o summary   & 78.9 & 87.3 & 90.6 & 94.4 & -2.9\\
    \midrule

    \parbox[t]{0.0mm}{\multirow{4}{*}{\rotatebox[origin=c]{90}{E5}}} 
    & \mname          & \textbf{87.0} & \textbf{92.8} & \textbf{94.5} & \textbf{96.5} & - \\
    & - w/o raw text  & 80.6 & 89.0 & 92.1 & 95.4 & -3.4 \\
    & - w/o keyword   & 84.6 & 91.3 & 93.3 & 96.0 & -1.4 \\
    & - w/o summary   & 83.9 & 90.3 & 92.8 & 95.5 & -2.1 \\
    \midrule

    \parbox[t]{0.0mm}{\multirow{4}{*}{\rotatebox[origin=c]{90}{BGE}}} 
    & \mname          & \textbf{83.7} & \textbf{90.6} & \textbf{93.0} & \textbf{95.3} & - \\
    & - w/o raw text  & 78.3 & 87.0 & 90.5 & 94.1 & -3.2 \\
    & - w/o keyword   & 81.0 & 89.0 & 91.3 & 94.3 & -1.8 \\
    & - w/o summary   & 79.7 & 88.1 & 91.1 & 94.2 & -2.4 \\
    \midrule

    \parbox[t]{0.0mm}{\multirow{4}{*}{\rotatebox[origin=c]{90}{GTE}}} 
    & \mname          & \textbf{84.0} & \textbf{90.8} & \textbf{93.1} & \textbf{96.0} & - \\
    & - w/o raw text  & 79.6 & 87.7 & 90.6 & 94.5 & -2.9 \\
    & - w/o keyword   & 81.8 & 89.2 & 91.8 & 94.7 & -1.6 \\
    & - w/o summary   & 80.4 & 88.5 & 91.4 & 94.5 & -2.3 \\
    
  \bottomrule
  \end{tabular}
  }
\vspace{-1em}
\caption{Ablation study of recall on WikiWeb2M, $\Delta$ refers to the average decrease of top 1.5, 3, 5, and 10.}
\vspace{-1.5em}
\label{tab:ablation_2m}
\end{table}

\subsection{Main Results} \label{ssec:main}
We display our main result in Table~\ref{tab:main} and summarise the our analysis with several key observations as follows:
\textbf{(1)} The size of chunk significantly impacts the recall. When comparing between different chunk size of FLC, we find out that larger chunks led to an increased recall. We believe that larger chunks are able to retain more information of the answer scope in a single chunk, which lead to better prediction from the retrieval. As shown in Table~\ref{tab:main}, the improvement from FLC-100 to FLC-300 is around 10-15\%.
\textbf{}(2) Content-aware chunking further reduces possibility of the ground truth answer scope being split. We define such error as the chunking error (see Section ~\ref{ssec:chunking}). As shown in Table~\ref{tab:main}, given that the chunk size of FLC and FLC-content approximately the same, we observe an improvement of 3-5\% in FLC-content.
\textbf{(3)} Each view of multi-view strategy tends to help retrieval achieves a higher recall than FLC.
Among each individual view, utilizing summary view generate the best results, while raw-text view generate the second best results. Despite keywords view down-performs overall due to text having poor semantic structure, we observe that keyword is able to solve some tasks which the other two view unable. This contributes to a positive impact (see Section ~\ref{ssec:ablation}).
\textbf{(4)} The multi-view strategy, which consolidates top-ranked results of raw-text, keywords, and summary views, can substantially improve the recall of all FLC and single-view baselines. We believe the improvement is mainly contributed by content-aware chunking and multi-view indexing. On the other hand, different views are able to rank the relevance of sections to question from different perspectives, thus providing complimentary information.

\begin{figure*} [t]
    \centering
    \begin{subfigure}[b]{0.76\columnwidth}
         \includegraphics[trim={0 0 2cm 0},clip, width=\linewidth]{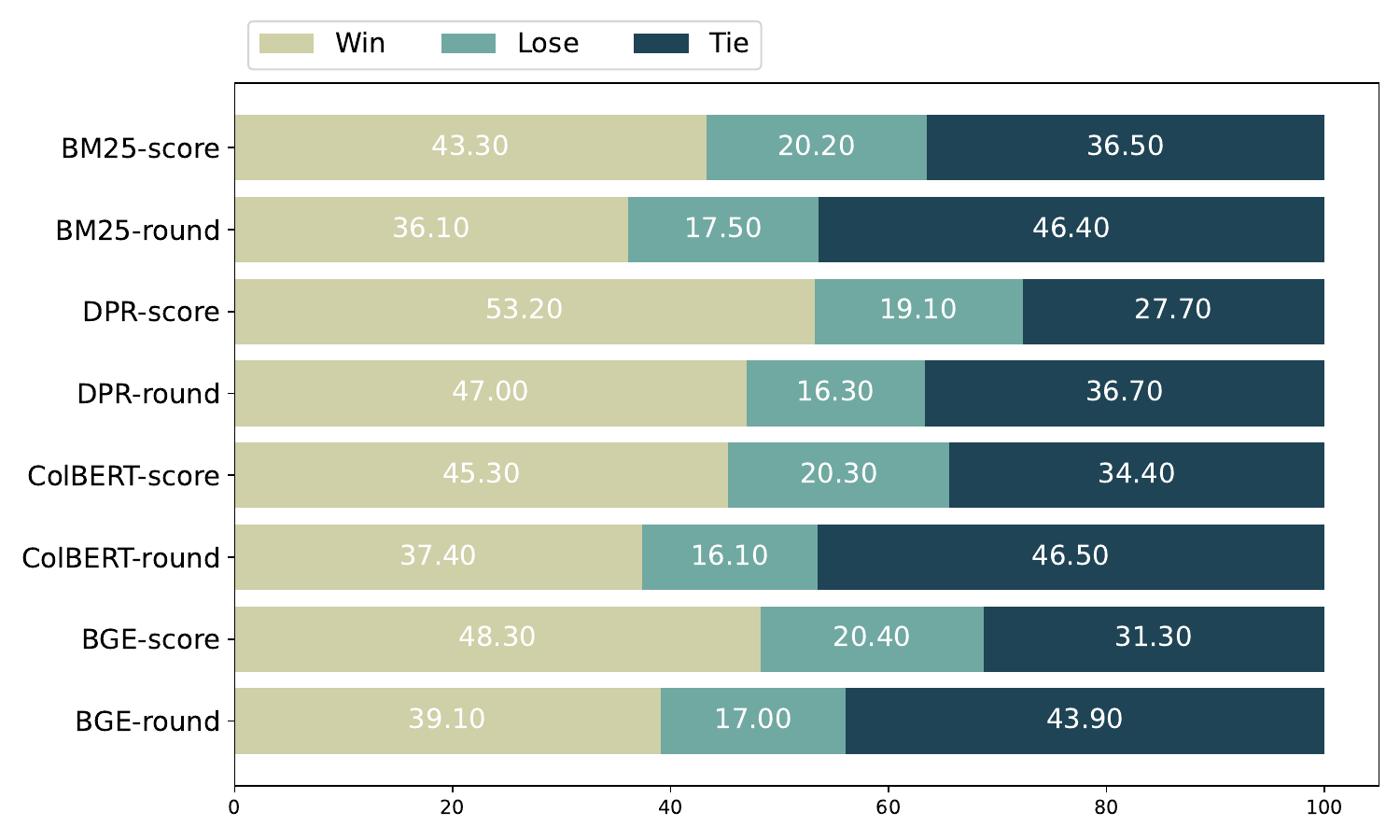}
         \vspace{-1.5em}
         \caption{Win, lose, tie rates for top $k=1.5$}
    \end{subfigure}
    ~
    \begin{subfigure}[b]{0.62\columnwidth}
         \includegraphics[trim={4.34cm 0 2cm 0},clip, width=\linewidth]{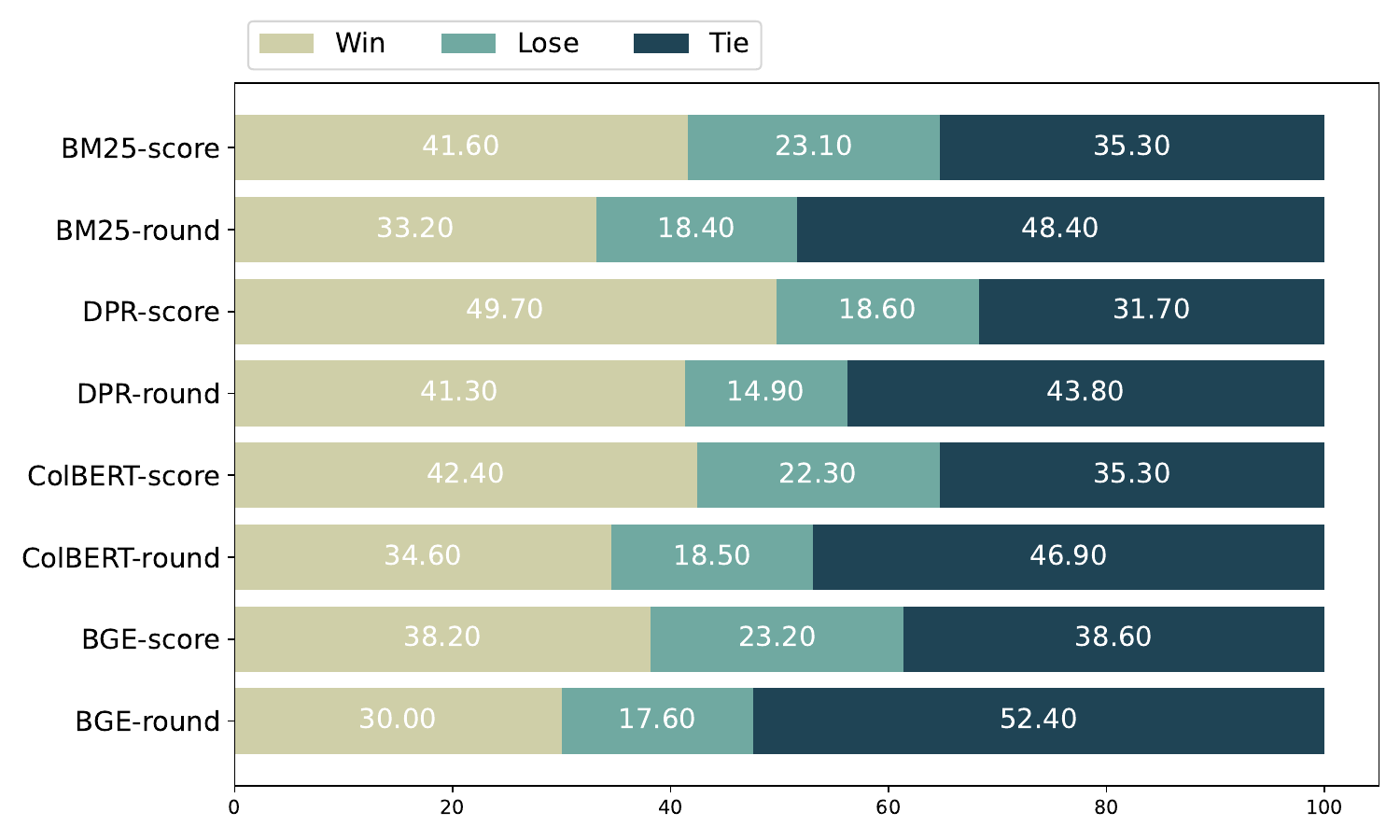}
         \vspace{-1.5em}
         \caption{Win, lose, tie rates for top $k=3$}
    \end{subfigure}
    ~
    \begin{subfigure}[b]{0.62\columnwidth}
         \includegraphics[trim={4.34cm 0 2cm 0},clip, width=\linewidth]{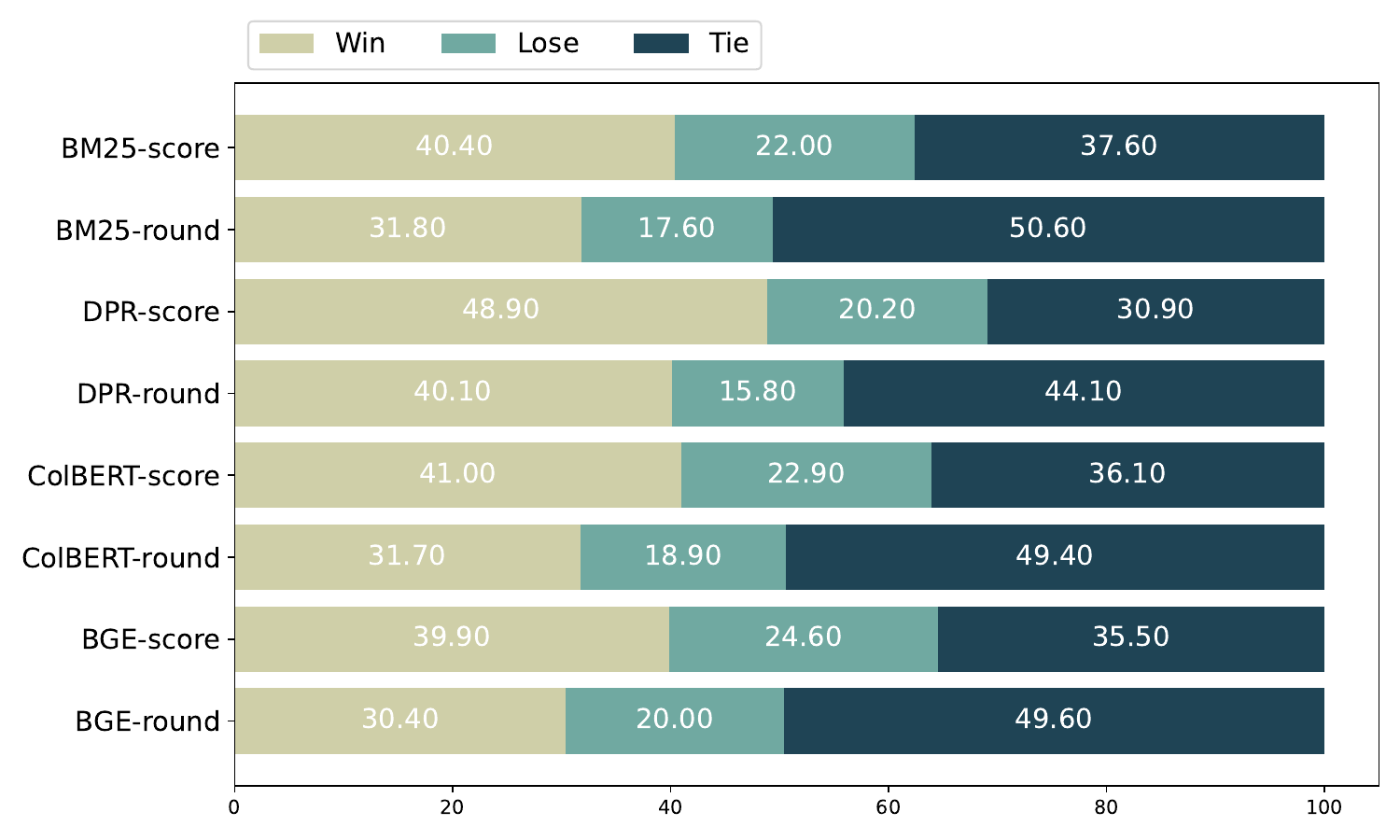}
         \vspace{-1.5em}
         \caption{Win, lose, tie rates for top $k=5$}
    \end{subfigure}
    \caption{The evaluation results of answer generation.}
    \label{fig:answer_eval}
\end{figure*}

\subsection{Ablation Study} \label{ssec:ablation}
We conducted an in-depth study by ablating each view from our multi-view indexing strategy and measuring the performance by recall.
From the results presented in Table~\ref{tab:ablation_2m}, we observe that:
\textbf{(1)} Removing the summary view leads to the most significant decrease in performance, ranging between 2 and 8\%. 
\textbf{(2)} Eliminating the raw-text view results in the second-most considerable performance drop, varying between 2 and 5\%.
\textbf{(3)} Disregarding the keywords view contributes to a decrease of performance ranging from 1 to 4\%. 

Thus, we infer that the impact of each view on the recall performance of retrieval, from the most to the least significant, is as follows: summary view, raw-text view, and keywords view. In conclusion, each view plays a crucial role in improving recall performance. More ablation results on NQ dataset are shown in Appendix~\ref{appendix:ablation}.

\subsection{Evaluation of Answer Generation}

We compare the performance of \mname against FLC-300 via the relevance of generated answers. 
For our experiments, we employ various retrieval methods, including BM25, DPR, ColBERT, and BGE. For each of \mname and FLC-300, we first use these retrievers to sample the sections related to the question. Given the retrieved sections, we proceed to generate answers using the prompt provided in Figure~\ref{fig:promp_ac}. The generated answers are then compared using pairwise comparison (see Section ~\ref{ssec:evaluation_metric}).

The results of this comparative assessment are displayed in Figure~\ref{fig:answer_eval}.
We find that \mname consistently demonstrates higher win rates than loss rates against FLC-300 across all retrievers and both evaluation metrics.

Positional bias in LLM may cause it to assign higher scores to the first answer in the prompt. Unlike score-based evaluation, which takes into account the magnitude of score differences, round-based evaluation is purely predicated on the number of rounds won by each answer. Consequently, we anticipate that the round-based evaluation will yield more ties than the score-based evaluation.

\subsection{Chunking Error} \label{ssec:chunking}
\begin{table}[t] 
\small
    \centering
  \begin{tabular}{l|l|ccc}
    \toprule
    \multirow{2}{*}{Chunk} & \multirow{2}{*}{Dataset}
    &  \multirow{2}{*}{FLC} & FLC- & Content \\
    & & & content & -aware\\
    \midrule
    \multirow{2}{*}{N=100}
    & Wiki-NQ & 66.4 & 50.8 & 0.0 \\
    & Wiki-2M & 75.3 & 60.9 & 0.0 \\
    \midrule

    \multirow{2}{*}{N=200} 
    & Wiki-NQ & 41.4 & 23.2 & 0.0 \\
    & Wiki-2M & 46.6 & 28.7 & 0.0 \\
    \midrule

    \multirow{2}{*}{N=300}
    & Wiki-NQ & 26.4 & 13.5 & 0.0 \\
    & Wiki-2M & 32.2 & 15.0 & 0.0 \\
    
  \bottomrule
  \end{tabular}
\caption{Chunking Error for each chunking method.}
\label{tab:chunk_error}
\end{table}
As previously discussed in Section~\ref{sec:introduction}, FLC tends to cause significant chunking errors. In this section, we elaborate the chunking errors from two fixed-length chunking strategies.

Firstly, the existing FLC method is content-agnostic. This is due to the fact the method divides the entire document into fixed-length chunks, which may inadvertently break a coherent section into separate parts.
Alternatively, we recommend a different FLC approach that segments each section of the document into fixed-length chunks. This would ensure that a chunk doesn't span across two different sections, thereby more robust to chunking errors.
In summary, our proposed content-aware chunking strategy ensures that no chunk extends over two sections, effectively reducing chunking errors. Results shown in Table~\ref{tab:chunk_error} highlight the impact of content-aware chunking on chunking error.

\section{Conclusion}
\label{sec:conclusion}
In this paper, we propose a new approach: \textbf{M}ulti-view \textbf{C}ontent-aware \textbf{indexing} (\textbf{MC-indexing}) for more effective retrieval of long document.
Specially, we propose a long document QA dataset which annotates not only the question-answer pair, but also the document structure and the document scope to answer this question.
We propose a content-aware chunking method to segment the document into content chunks according to its organizational content structure.
We design a multi-view indexing method to represent each content chunk in raw-text, keywords, and summary views.
Through extensive experiments, we demonstrate that content-aware chunking can eliminate chunking errors, and multi-view indexing can significantly benefit long DocQA.
For future work, we would like to explore how to use the hierarchical document structure for more effective retrieval. Moreover, we would like to train or finetune a retriever that can generate more fine-grained or nuanced embeddings across multiple views.

\clearpage
\section*{Limitations}
\label{sec:limit}
The limitations of our method \mname, can be evaluated from two primary perspectives.

Firstly, our method heavily depends on the structured format of a document. When the document lacks clear indications of content structure, applying our content-aware chunking technique becomes challenging. However, we would like to emphasize that our work focuses on structured indexing and retrieval of long documents, and long documents usually have structured content to be utilised. It is unusual to encounter lengthy and poorly structured documents in which the authors have written tens of thousands of words without providing clear document section or chapter demarcations. 

To expand the usability of our method to unstructured documents, we propose the development of a \emph{section chunker} that can segment documents without predefined content structures into atomic and coherent semantic units as a future research direction.

Secondly, short documents, being within the Large Language Model's (LLM) capacity, which means structured layout might not be required for the model to perform Question Answering (QA) tasks efficiently. Hence, we clarify that our method does not aim to enhance retrieval performance on unstructured short document. In contrast, our method can significantly benefit the retrieval of structured long documents.


\bibliography{custom}

\appendix
\section{Appendix}
\label{sec:appendix}

\subsection{Implementation Details of Retrievers} \label{appendix:implementation_retriever}

\begin{table*}[t]
\small
    \centering
    \resizebox{\linewidth}{!}{%
\begin{tabular}{|l|l|l|l|}
\hline
\textbf{Model} & \textbf{Dimension} & \textbf{Base Model} & \textbf{HuggingFace Checkpoint}                               \\ \hline
DPR &
  768 &
  bert-base &
  \begin{tabular}[c]{@{}l@{}}\url{https://huggingface.co/facebook/dpr-ctx\_encoder-multiset-base}\\ \url{https://huggingface.co/facebook/dpr-question\_encoder-multiset-base}\end{tabular} \\ \hline
ColBERT    & 768 & bert-base  & \url{https://huggingface.co/colbert-ir/colbertv2.0}      \\ \hline
Contriever & 768                 & bert-base  & \url{https://huggingface.co/facebook/contriever-msmarco} \\ \hline
E5         & 1024                & bert-large & \url{https://huggingface.co/intfloat/e5-large-v2}        \\ \hline
BGE        & 1024                & RetroMAE   & \url{https://huggingface.co/BAAI/bge-large-en-v1.5}      \\ \hline
GTE        & 1024                & bert-large & \url{https://huggingface.co/thenlper/gte-large}          \\ \hline
\end{tabular}}
\caption{Implementation details for Dense Models}
\label{tab:dense-implementation-details}
\end{table*}

We used 6 dense retrieval models and 2 sparse retrieval models in section~\ref{sec:experiment}. The dense retrieval models deployed are namely DPR (Dense Passage Retriever), ColBERT, Contriever, E5, BGE and GTE. These models use the WordPiece tokenizer from Bert and also inherit the maximum input length of 512 tokens from Bert. We use pre-trained checkpoints available on HuggingFace \footnote{\url{https://huggingface.co/}}; the specific checkpoint information can be found in Table~\ref{tab:dense-implementation-details} alongside other configuration details. Additionally, we make use of the sentence-transformer library\footnote{\url{https://www.sbert.net/}} when deploying E5, BGE and GTE.

The 2 sparse retrieval models implemented are BM25 and TF-IDF (Term Frequency - Inverse Document Frequency). Note that when calculating scores for BM25 and TF-IDF for each question, we restrict the set of corpus to chunks appearing in the sole relevant Wikipedia article. For BM25, we use the code from github repository \url{https://github.com/dorianbrown/rank\_bm25}. For TF-IDF we use the TF-IDF Vectorizer from scikit-learn library \footnote{\url{https://scikit-learn.org/stable/modules/generated/sklearn.feature\_extraction.text.TfidfVectorizer.html}}. We briefly describe how we rank document using the TF-IDF vectorizer here. First, given the corpus (i.e. the chunks appearing in the sole relevant Wikipedia article) we convert each chunk into a sparse vector with each entry indicating the TF-IDF score of each word appearing in the chunk. Next, we convert the question into a sparse vector. Finally to rank each chunk, we calculate the cosine similarity between the question sparse vector and sparse vectors of each individual chunk.

\begin{table}[t] 
\small
    \centering
    \resizebox{\linewidth}{!}{%
  \begin{tabular}{ll|cccc|c}
    \toprule
    \multicolumn{2}{c|}{\multirow{1}{*}{Chunk Scheme}}
    &  Top1.5 & Top3 & Top5 & Top10 & $\Delta$ \\
    \midrule
    \parbox[t]{0.0mm}{\multirow{4}{*}{\rotatebox[origin=c]{90}{TF-IDF}}} 
    & \mname         & 40.9 & 54.1 & 67.6 & 85.7 & - \\
    & - w/o raw text & 32.4 & 49.5 & 63.5 & 83.8 & -4.8\\
    & - w/o keyword  & 34.5 & 51.2 & 64.5 & 84.3 & -3.4\\
    & - w/o summary  & 32.4 & 47.6 & 60.1 & 82.6 & -6.4\\
    \midrule

    \parbox[t]{0.0mm}{\multirow{4}{*}{\rotatebox[origin=c]{90}{BM25}}} 
    & \mname          & 36.9 & 47.6 & 60.1 & 78.2 & - \\
    & - w/o raw text  & 25.9 & 41.6 & 52.0 & 72.9 & -7.6 \\
    & - w/o keyword   & 30.4 & 43.2 & 55.1 & 74.2 & -5.0 \\
    & - w/o summary   & 27.6 & 41.6 & 54.4 & 72.7 & -6.6 \\
    \midrule

    \parbox[t]{0.0mm}{\multirow{4}{*}{\rotatebox[origin=c]{90}{DPR}}} 
    & \mname          & 58.4 & 75.1 & 87.5 & 95.0 & - \\
    & - w/o raw text  & 53.1 & 71.0 & 81.7 & 93.5 & -4.2\\
    & - w/o keyword   & 52.7 & 71.2 & 82.6 & 93.3 & -4.0\\
    & - w/o summary   & 49.8 & 69.1 & 81.2 & 90.5 & -6.4\\
    \midrule

    \parbox[t]{0.0mm}{\multirow{4}{*}{\rotatebox[origin=c]{90}{ColBERT}}} 
    & \mname          & 62.3 & 77.1 & 85.2 & 94.8 & - \\
    & - w/o raw text  & 54.8 & 71.7 & 81.4 & 93.5 & -4.5 \\
    & - w/o keyword   & 55.8 & 72.5 & 81.1 & 93.7 & -4.1 \\
    & - w/o summary   & 55.6 & 72.4 & 81.2 & 93.2 & -4.2 \\
    \midrule

    \parbox[t]{0.0mm}{\multirow{4}{*}{\rotatebox[origin=c]{90}{Contriever}}} 
    & \mname          & 52.2 & 70.8 & 82.1 & 92.7 & - \\
    & - w/o raw text  & 46.9 & 65.5 & 79.4 & 89.2 & -4.2 \\
    & - w/o keyword   & 46.1 & 64.7 & 78.5 & 88.7 & -4.9 \\
    & - w/o summary   & 45.1 & 65.0 & 77.6 & 91.6 & -4.6\\
    \midrule

    \parbox[t]{0.0mm}{\multirow{4}{*}{\rotatebox[origin=c]{90}{E5}}} 
    & \mname          & 69.6 & 85.3 & 91.8 & 97.2 & - \\
    & - w/o raw text  & 63.3 & 81.4 & 90.3 & 95.9 & -3.2 \\
    & - w/o keyword   & 62.8 & 80.0 & 91.3 & 96.4 & -3.3 \\
    & - w/o summary   & 60.9 & 80.3 & 91.1 & 96.7 & -3.7 \\
    \midrule

    \parbox[t]{0.0mm}{\multirow{4}{*}{\rotatebox[origin=c]{90}{BGE}}} 
    & \mname          & 63.1 & 78.8 & 89.2 & 95.4 & - \\
    & - w/o raw text  & 58.0 & 74.9 & 86.2 & 94.0 & -3.3 \\
    & - w/o keyword   & 57.5 & 73.7 & 85.7 & 94.9 & -3.7 \\
    & - w/o summary   & 56.7 & 74.4 & 85.8 & 94.4 & -3.8 \\
    \midrule

    \parbox[t]{0.0mm}{\multirow{4}{*}{\rotatebox[origin=c]{90}{GTE}}} 
    & \mname          & 62.3 & 77.8 & 88.0 & 95.4 & - \\
    & - w/o raw text  & 55.5 & 73.0 & 85.8 & 94.5 & -3.7 \\
    & - w/o keyword   & 57.3 & 74.7 & 86.1 & 94.8 & -2.7 \\
    & - w/o summary   & 57.7 & 74.0 & 85.0 & 94.0 & -3.2 \\
    
  \bottomrule
  \end{tabular}
  }
\caption{Ablation study of recall on NQ, $\Delta$ refers to the average decrease of top 1.5, 3, 5, and 10.}
\label{tab:ablation_nq}
\end{table}

\subsection{Extended Ablation Study on NQ}
\label{appendix:ablation}
In this section, we reported the ablation results of \mname on NQ dataset, serving as the extension of Section~\ref{ssec:ablation}.
From the data in Table~\ref{tab:ablation_nq}, it's evident that:
\textbf{(1)} Removing the raw-text view leads to the most significant performance drop, ranging between 3.2 and 7.6\%.
\textbf{(2)} Eliminating the summary view results in the second-most considerable performance drop, varying between 3.2 and 6.6\%.
\textbf{(3)} Disregarding the keywords view contributes to a performance drop between 2.7 and 5\%.

\subsection{Question Generation for WikiWeb2M}
\label{ssec:q_type}
\begin{figure}
    \centering
    \includegraphics[width=0.9\columnwidth]{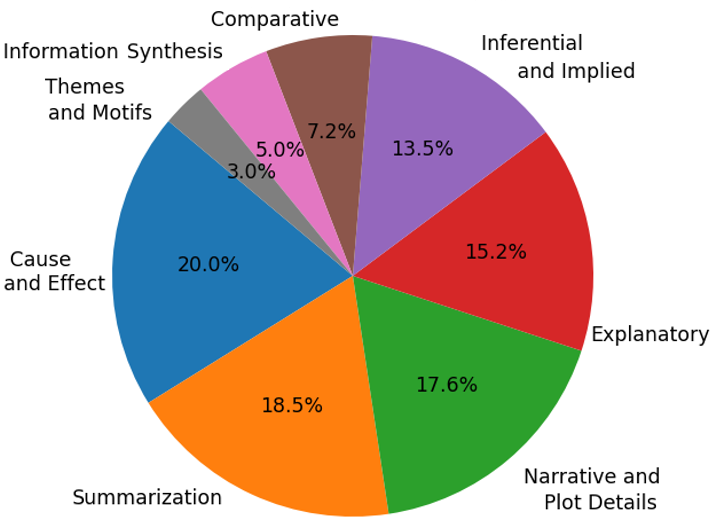}
    \caption{Pie chart of question type distribution.}
    \label{fig:q_distributation}
\end{figure}
We aim to generate question that tends to rely on a long answer scope. Typically, the length of answer scope ranges from 50 to 500 tokens.
We define questions of the following 8 types:
\begin{itemize}[leftmargin=*, itemsep=-0.35em, topsep=-0.0em]
    \item \textit{Narrative and Plot Details}: inquire specific details or sequence of events in a narrative (\eg a story, movie, or historical account).
    \item \textit{Summarization}: require the summarization of a long passage, argument, or complicated process.
    \item \textit{Inferential and Implied}: depend on understanding subtleties and reading across a long passage.
    \item \textit{Information Synthesis}: inquire the synthesis of information dispersed across a long passage.
    \item \textit{Cause and Effect}: understand the causal relationship between events in a long passage.
    \item \textit{Comparative}: ask for comparisons between different ideas, characters, or events within a text.
    \item \textit{Explanatory}: ask for explanations of complex concepts or processes that are described in detail. 
    \item \textit{Themes and Motifs}: consider entire text to identify patterns and conclude on central messages.
\end{itemize}
The distribution of generated question types is shown in Figure~\ref{fig:q_distributation}.

\subsection{Question Answer Annotation for WikiWeb2M}
\label{ssec:annotation}
For each given section, we request the advanced commercial LLM to generate 3 questions, the corresponding answers and identify the raw text that maps to the answer. In our prompt from Figure ~\ref{fig:prompt_query}, we provide LLM the raw text of the given section, the description of the 8 question types from Appendix ~\ref{ssec:q_type} and our designed prompt instruction. Our prompt instruction ensures LLM to generate the continuous context sentences to sufficiently answer the question. The answer scope is then used to evaluate the retrieval efficiency of MC-indexing.

\subsection{Prompt Design} \label{appendix:prompt}
In this paper, we utilize the following prompts on the advanced commercial LLM to facilitate the respective process:
\begin{itemize}[leftmargin=*, itemsep=-0.0em, topsep=-0.0em]
    \item The generation of WikiWeb2M question, question type, answer, and answer contextual sentences. The prompt is shown in Figure~\ref{fig:prompt_query}.
    \item The contextual sentences retrieval when provided with a long document or a section of the document. This is used to evaluate if the advanced commercial LLMs can directly cope with long document. The prompt is shown in Figure~\ref{fig:promp_qa}.
    \item The generation of summary for the sections consisting of more than 200 tokens. The generated summary is used as additional view for document indexing. The prompt is shown in Figure~\ref{fig:promp_summary}.
    \item The generation of the list of keywords for each section. The generated keywords list is used as additional view for document indexing. The prompt is shown in Figure~\ref{fig:promp_keywords}.
    \item The answer generation when provided with retrieved top $k$ chunks or sections. The prompt is shown in Figure~\ref{fig:promp_ac}.
    \item The automatic answer evaluation of two answers, given the ground truth answer. This is used to evaluate the answer quality. This prompt is shown in Figure~\ref{fig:promp_score}.
\end{itemize}

\subsection{Evaluating LLM on Long Document QA.}
\label{ssec:gpt_long_context}

We assess if state-of-the-art LLMs can effectively handle long document Question-Answering (QA).
We have opted for the Span-QA setting to simplify the process. Here, the gold answer is a span of raw text extracted directly from the input document. We then measure the precision, recall, and F$_1$ score of the retrieved span against this gold answer.

Since the most advanced commercial LLM is capable of taking fairly long input, it is tasked to analyze longer documents, approximately 30k tokens, and given 2,000 questions to answer. These questions are all sourced from our Wiki-2M dataset (in Section~\ref{ssec:data_wiki2m}).
To further determine if LLMs can perform more proficiently given a shorter answer scope, we adapt our input approach. Instead of feeding the model the entire document, only the corresponding section is inputted, which contains an average of 370 tokens. 

Finally, we compare the performances when using the entire document versus using only a section. This allows for a comprehensive evaluation of the models' capabilities in handling different lengths of text. The result is elaborated in Figure~\ref{fig:llm_long_context}.

\begin{figure*}[!b]
\centering
\begin{minipage}{\textwidth}
{\footnotesize
\begin{verbatim}
You are a sophisticated question generator. You need to use the reference text to generate a question, 
with its question type, and the supporting context sentences, and the short answer. 

The generation should strictly follow the following guidelines:
(1) The question must be sufficiently answered by the reference text only; 
(2) The question need to be short and accurate;
(3) All supporting context sentences must be the original text from the reference text;
(4) The question should need long context (more than 5 sentences) to answer accurately;
(5) The type of each question needs to be ONE from the following eight types:
1. **Questions about Narrative and Plot Details**: inquire about specific details or the sequence of events 
    in a narrative (such as a story, movie, or historical account) require understanding the entire context 
   to provide an accurate answer.
2. **Summarization Questions**: require the summarization of a long passage, argument, or a complicated 
    process rely on understanding the full context to capture the essence of the content without omitting 
   crucial details.
3. **Inferential and Implied Questions**: depend on understanding subtleties and reading between the lines. 
   They may involve inferring the author's intent, the mood of the characters in a story, or the 
   implications of certain actions, which can't be answered with a direct quote from the text.
4. **Questions Requiring Synthesis of Information**: necessitate the synthesis of information dispersed
   across a long passage or multiple passages, requiring an understanding of the broader context to 
   answer correctly.
5. **Cause and Effect Questions**: to understand the causal relationship between events in a text, one 
   often needs to consider a substantial portion of the context to identify the factors that led to 
   a particular outcome.
6. **Comparative Questions**: ask for comparisons between different ideas, characters, or events within
   a text often require a comprehensive understanding of each element being compared.
7. **Explanatory Questions**: ask for explanations of complex concepts or processes that are described
   in detail within the text. Answering these questions accurately requires a deep understanding of the 
   entire explanation as presented.
8. **Questions about Themes and Motifs**: when asked about the overarching themes or motifs in a text, one
   must consider the entire work to identify patterns and draw conclusions about the central messages.

**Reference text**:
$text

Return the question and answer in the following json format:
{question:"...", type:"...", answer:"...", answer_context:"..."} 
\end{verbatim}}
\end{minipage}
\caption{Prompt used for question and answer generation.}
\label{fig:prompt_query}
\end{figure*}
\begin{figure*}[!b]
\centering
\begin{minipage}{\textwidth}
{\footnotesize
\begin{verbatim}
You are helpful question answering assistant. Given a question and the reference text, you need to find 
sufficient context to answer this question. The context sentences must be the original text of reference 
text. Note that you must not answer these question. 

**Question**: $question

**Reference Text**: $reference

Return the result in json format: {"context": ..., "}
\end{verbatim}}
\end{minipage}
\caption{Prompt template designed to find the relevant answer scope given the question and section text.}
\label{fig:promp_qa}
\end{figure*}
\begin{figure*}[!b]
\centering
\begin{minipage}{\textwidth}
{\footnotesize
\begin{verbatim}
You are a helpful summarization assistant. Please help me summarize the following section into no more
than 10 sentences or 200 words.

**Section Name**:
$section_name

**Section Text**:
$section_text

\end{verbatim}}
\end{minipage}
\caption{Prompt template designed to provide summary for section given its corresponding name and text.}
\label{fig:promp_summary}
\end{figure*}
\begin{figure*}[!b]
\centering
\begin{minipage}{\textwidth}
{\footnotesize
\begin{verbatim}
You are a helpful keyword extractor. You need to extract keywords from the following section. The keywords 
should consist of concepts, entities, or important descriptions that are related to the section text, which 
could be used to answer any questions from users.

**Section Name**:
$section_name

**Section Text**:
**Beginning of text**
$section_text$
**End of text**

Please output format in list format: [...]. Do not output anything else aside from this list.

\end{verbatim}}
\end{minipage}
\caption{Prompt template designed to provide keywords for section given its corresponding name and text.}
\label{fig:promp_keywords}
\end{figure*}
\begin{figure*}[!b]
\centering
\begin{minipage}{\textwidth}
{\footnotesize
\begin{verbatim}
You are a helpful question answering assistant. You are good at answering question based on provided contents.

**Contents**: $quotes

**Question**: $question

**Instruction:**
Assume you do not have any background and internal knowledge about this given contents and question. 
You need to answer the question using the given contents only. The answer need to be short and accurate.

\end{verbatim}}
\end{minipage}
\caption{Prompt template designed to answer question based on the retrieved results.}
\label{fig:promp_ac}
\end{figure*}
\begin{figure*}[!b]
\centering
\begin{minipage}{\textwidth}
{\footnotesize
\begin{verbatim}
You are a helpful assistant for evaluating answers. Given a question and ground truth answer, there will be 
two possible answers. Provide a score from 0-10 for each answer.

**Question**: $question

**Ground truth answer**: $ground_truth_answer

**Answer 1**: $answer_1
**Answer 2**: $answer_2


**Instruction:**
Assume you do not have any background and internal knowledge about this given contents and question. You 
need to evaluate each answer and give a score based on the ground truth answer. 
You must write out your reasoning of the score based on relevance to the answer. If both answers are 
exactly similar, you must ensure the scores and reasoning for both answers are the same.
Finally in a new line, you must return the scores and nothing else. The scores must be returned in the 
following json format:
{"answer_1_score":"...", "answer_2_score":"..."}

\end{verbatim}}
\end{minipage}
\caption{Prompt template designed to provide score for each answer in pair-wise evaluations.}
\label{fig:promp_score}
\end{figure*}

\end{document}